\documentclass[letterpaper, 10 pt, conference]{ieeeconf}  

\IEEEoverridecommandlockouts

\usepackage{hyperref}       
\usepackage{url}            

\usepackage{booktabs}       

\usepackage{arydshln}
\usepackage{multirow}

\usepackage{amsfonts}       
\usepackage{nicefrac}       
\usepackage{microtype}      
\usepackage{xcolor}         

\usepackage{fancyhdr,amsmath,xspace,algcompatible, enumerate, paralist, mathtools, tabularx, hhline,xcolor,colortbl,tabu,forest,xcolor,array,multirow}
\usepackage[noline, ruled,boxed]{algorithm2e}
\usepackage[capitalise]{cleveref}
\usepackage[export]{adjustbox}
\usepackage{pgf,tikz}
\usepackage{graphicx}
\usepackage{epstopdf}
\usepackage{stmaryrd}
\usepackage{subcaption}
\usepackage{bm}
\usepackage{mathrsfs}
\usepackage{type1cm}
\usepackage{csquotes}
\usepackage{amssymb}
\usepackage{comment}
\usepackage{multirow}
\usepackage{pifont}
\usepackage{verbatim}
\usepackage{float}
\usepackage{wrapfig}
\usepackage{thmtools}
\usepackage{dsfont}
\usepackage{thm-restate}
\usepackage{soul}
\usepackage{comment}
\usepackage{diagbox}

\makeatletter
\newcommand{\thickhline}{%
    \noalign {\ifnum 0=`}\fi \hrule height 1pt
    \futurelet \reserved@a \@xhline
}
\newcolumntype{"}{@{\hskip\tabcolsep\vrule width 1pt\hskip\tabcolsep}}
\makeatother

\title{\LARGE \bf
From Twitter to Reasoner: Understand Mobility Travel Modes and Sentiment Using Large Language Models
}

\author{Kangrui Ruan$^{1}$, Xinyang Wang$^{2}$, Xuan Di$^{3,*}$ 
\thanks{$^{*}$Corresponding author: Xuan Di.}
\thanks{$^{1}$Kangrui Ruan is with the Department of Civil Engineering and Engineering Mechanics (CEEM), Columbia University, New York, NY, 10027, USA
        {\tt\small (E-mail: kr2910@columbia.edu)}}%
\thanks{$^{2}$Xinyang Wang is with the Data Science Institute (DSI), Columbia University, New York, NY, 10027 USA
        {\tt\small (E-mail: xw2964@columbia.edu)}}%
\thanks{$^{3}$Xuan Di is with the Department of CEEM, and also with the DSI, Columbia University, New York, NY, 10027 USA {\tt\small (E-mail: sharon.di@columbia.edu)}.}
}

\begin{document}

\maketitle
\thispagestyle{empty}
\pagestyle{empty}

\begin{abstract}
Social media has become an important platform for people to express their opinions towards transportation services and infrastructure, which holds the potential for researchers to gain a deeper understanding of individuals' travel choices, for transportation operators to improve service quality, and for policymakers to regulate mobility services.
A significant challenge, however, lies in the unstructured nature of social media data. In other words, textual data like social media is not labeled, and large-scale manual annotations are cost-prohibitive.
In this study, we introduce a novel methodological framework utilizing Large Language Models (LLMs) to infer the mentioned travel modes from social media posts, and reason people's attitudes toward the associated travel mode, without the need for manual annotation. 
We compare different LLMs along with various prompting engineering methods in light of human assessment and LLM verification.
We find that most social media posts manifest negative rather than positive sentiments.
We thus identify the contributing factors to these negative posts and, accordingly, propose recommendations to traffic operators and policymakers.
\end{abstract}


\vspace{-2.0 mm}
\section{INTRODUCTION}
\vspace{-1.0 mm}


Social media significantly influences our daily lives, with approximately two-thirds of American adults visiting social networks regularly \cite{perrin2015social}. 
This widespread utilization positions social media as a vital source for the acquisition and dissemination of current information, highlighting its growing appeal as a cost-effective alternative to traditional data collection methods \cite{rashidi2017exploring}.
As a result, social media has evolved from a simple platform for connecting individuals to an extensive and invaluable repository of data. 
This evolution facilitates the study of human interactions and behaviors across various fields, e.g., transportation \cite{yao2021twitter}, emergency management \cite{panagiotopoulos2016social}, and so on.

\vspace{-0.5 mm}
\subsection{Related Work}
\vspace{-1. mm}

Previously, numerous studies have utilized social media data for transportation research, including the classification of urban activity patterns \cite{hasan2014urban}, estimation of travel activity spaces \cite{lee2016activity}, examination of longitudinal travel behavior \cite{zhang2017potentials}, incidents detection \cite{gu2016twitter}, and so on.
Specifically, the authors in \cite{hasan2014urban} utilize Latent Dirichlet Allocation to classify individual activity patterns.
\cite{lee2016activity} estimates the differences between weekday and weekend activity spaces through geo-tagged tweets from different users.
Gu et al. \cite{gu2016twitter} first manually annotate tweets relevant to the traffic incidents and develop a Semi-Naive-Bayes model to classify the results. 
Similarly, Chen et al. \cite{chen2023sentiment} also manually label a small amount of tweets and train two separate classifiers for different objectives. 
Ye et al. \cite{ye2021impact} explore Twitter data to understand attitude changes in travel behaviors during the COVID-19 pandemic.

However, there are several potential challenges identified in the previous research: (1) Manual annotation of the large volume of unlabeled tweets is extremely expensive and time-intensive \cite{brown2020language}. (2) When various objectives exist, e.g., in analyzing travel modes vs. sentiments, previous researchers might need to develop distinct models \cite{chen2023sentiment,li-19-segmentation}, with the risk of cascading errors passed from travel mode classification to sentiment analysis. (3) The accessibility of geo-location information is frequently limited, as many users opt not to share their precise location coordinates \cite{luceri2020measurement}. Based on \cite{ruan2022learning,ruan2023causal,ruan2024causal}, such unobserved variables might pose substantial risks on learning processes.

To tackle these challenges, we propose a novel framework purely based on text, and leverage Large Language Models (LLMs), utilizing their exceptional performance \cite{wei2021finetuned}. 
LLMs are recently popular and have already been applied in various fields, e.g., computer vision \cite{dosovitskiy2020image}, agriculture \cite{lin2023mmst}, voice assistants \cite{ruan2024s2e,chen2024neural} and so on.
Without the need for manual annotation, the proposed framework can simultaneously predict travel modes, sentiments, and summarize the reasons. Therefore, this approach obviates the need for different classifiers designed for distinct objectives, streamlining the analytical process.

\begin{figure*}[!ht]
\centering
  \includegraphics[width = 0.92 \linewidth]{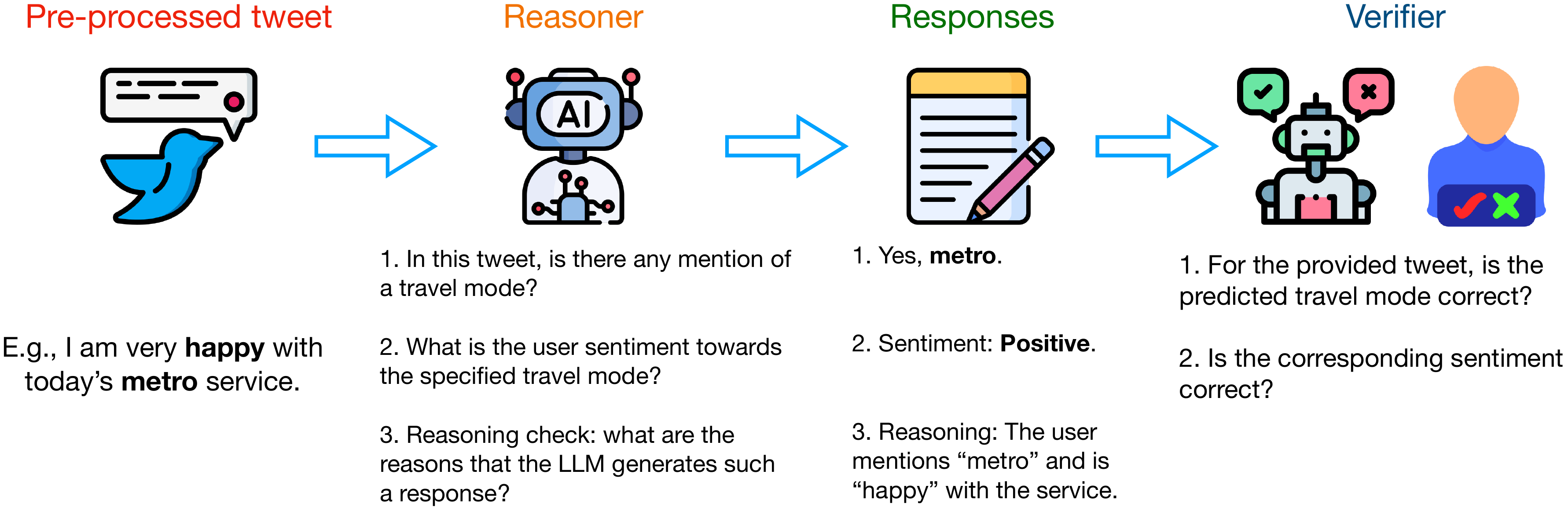}
  \vspace{-1. mm}
  \caption{The overall structure for the proposed framework. For each pre-processed tweet, the reasoner predicts the travel mode and corresponding sentiment while also performing a reasoning check. Subsequently, the verifier reviews and confirms the validity of the generated responses.}
  \vspace{-6. mm}
  \label{fig:overall_structure}
\end{figure*}

\vspace{-0.5 mm}
\subsection{Contributions of this paper}
\vspace{-1. mm}
The contributions of this paper can be summarized as follows: (1) We propose a novel pure-text-based framework based on Large Language Models (LLMs) to analyze social media data. 
This approach effectively infers the travel modes mentioned within the collected tweets and facilitates an in-depth understanding of public attitudes and concerns regarding various travel modes.
(2) We conduct a comparative analysis for different LLMs and different prompting engineering methods to determine their efficacy in understanding travel modes and the corresponding sentiment. The proposed framework is validated through systematic human evaluations and LLM verifications. (3) Based on the identified travel modes and public attitudes, we delve into the underlying reasons for negative attitudes and offer several specific recommendations for potential policy adjustments.

The structure of the paper is organized as follows. \Cref{sec:data_collection} details how the social media data is collected and used for analysis. 
\Cref{sec:methodology} elaborates on the developed framework, specifically focusing on LLMs and prompting engineering techniques. 
\Cref{sec:results} presents the major results, including  comparisons evaluated by both human assessments and LLM, analyses of travel modes and primary causes of dissatisfaction.
\Cref{sec:conclusion} concludes the paper.

\vspace{-2.0 mm}
\section{Data Collection} \label{sec:data_collection}
\vspace{-1.0 mm}

In this section, we discuss how the dataset is collected from Twitter.
We systematically gather tweets related to different travel modes using a structured list of keywords. 
For instance, to collect potentially relevant tweets about subways, we utilize keywords such as ``subway", ``metro", ``path", ``MTA", ``LIRR", ``train", ``light rail", ``transit" and so on.
Similarly, for buses, the keywords include ``bus" and ``public transport."
Although these tweets are collected based on specific keywords, there is no guarantee that they pertain to any particular travel mode. For example, the keyword ``subway" could refer to either the metro system or the restaurant. 
In other words, \textbf{the collected tweets are unlabelled}.
Therefore, when there is not any specific mode mentioned, we designate the travel mode field as `NA'.

Generally speaking, the collected data covers periods from 2020 to 2022, and is geographically mainly focused on New York City (NYC), because of its diverse travel modes.
Additionally, the dataset consists of more than 250,000 tweets, each record containing the user ID, username, tweet ID, timestamp, the corresponding text, and so on. 
Due to privacy considerations, this study focuses mainly on the textual content of the tweets themselves rather than the demographic information of the users, such as gender or age.

\begin{figure*}[!th]
\centering
\resizebox{0.92\linewidth}{!}{
\begin{subfigure}[b]{1\columnwidth}\centering
    \includegraphics[width=\linewidth]{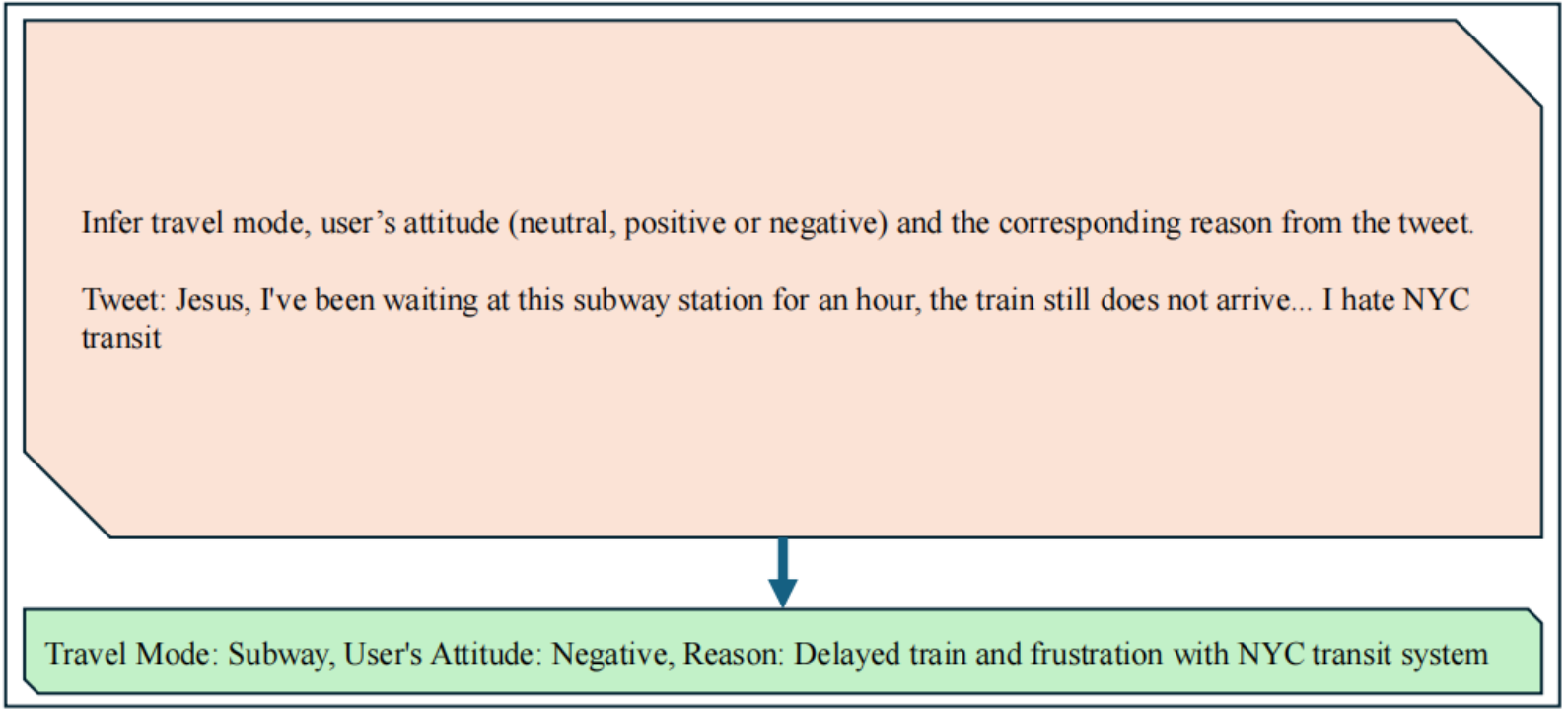}
    \caption{Instruction-following}
\label{fig:prompting:zero}
\end{subfigure}
\hfill
\begin{subfigure}[b]{1\columnwidth}\centering
    \includegraphics[width=\linewidth]{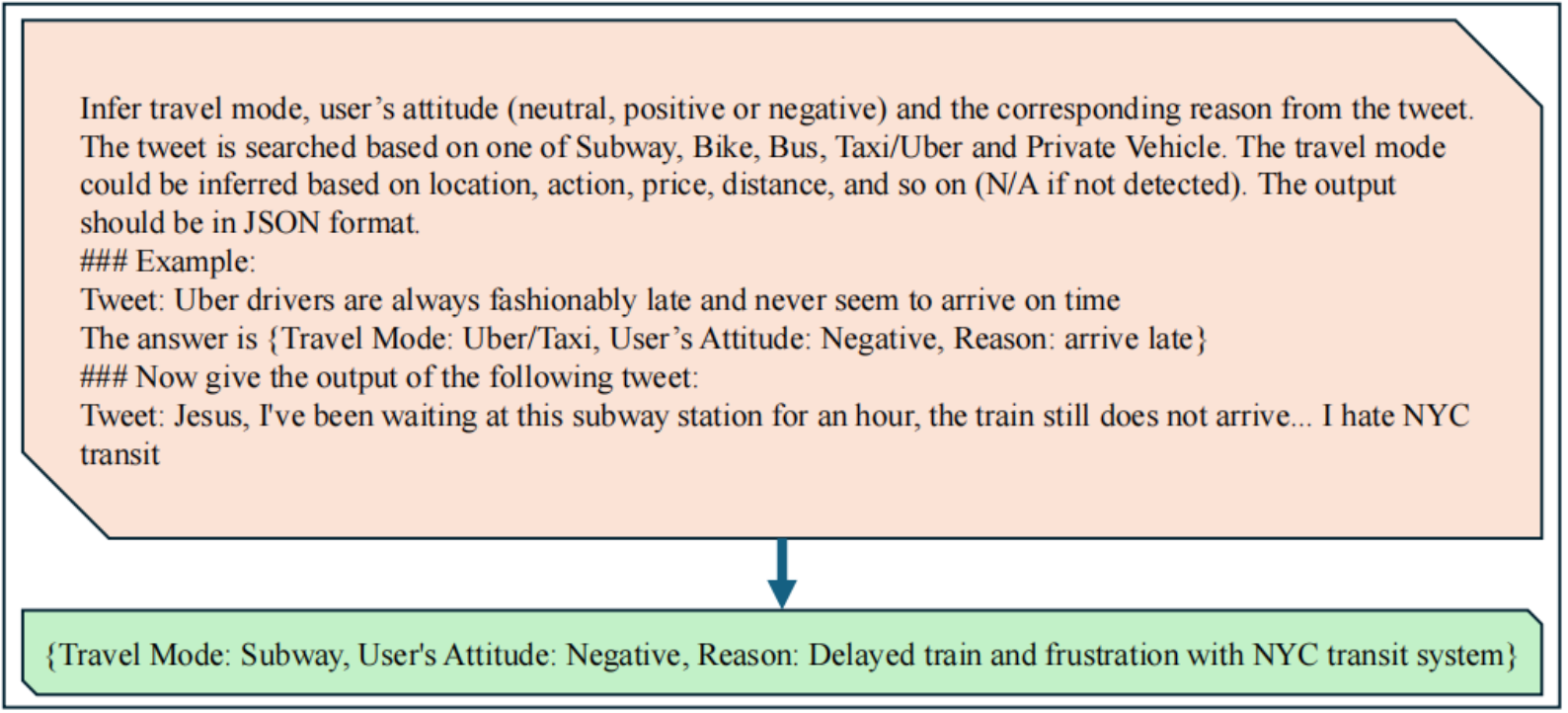}
    \caption{In-Context Learning (ICL)}
\label{fig:prompting:icl}
\end{subfigure}
}
\centering
\resizebox{0.92\linewidth}{!}{
\begin{subfigure}[b]{1\columnwidth}\centering
    \includegraphics[width=\linewidth]{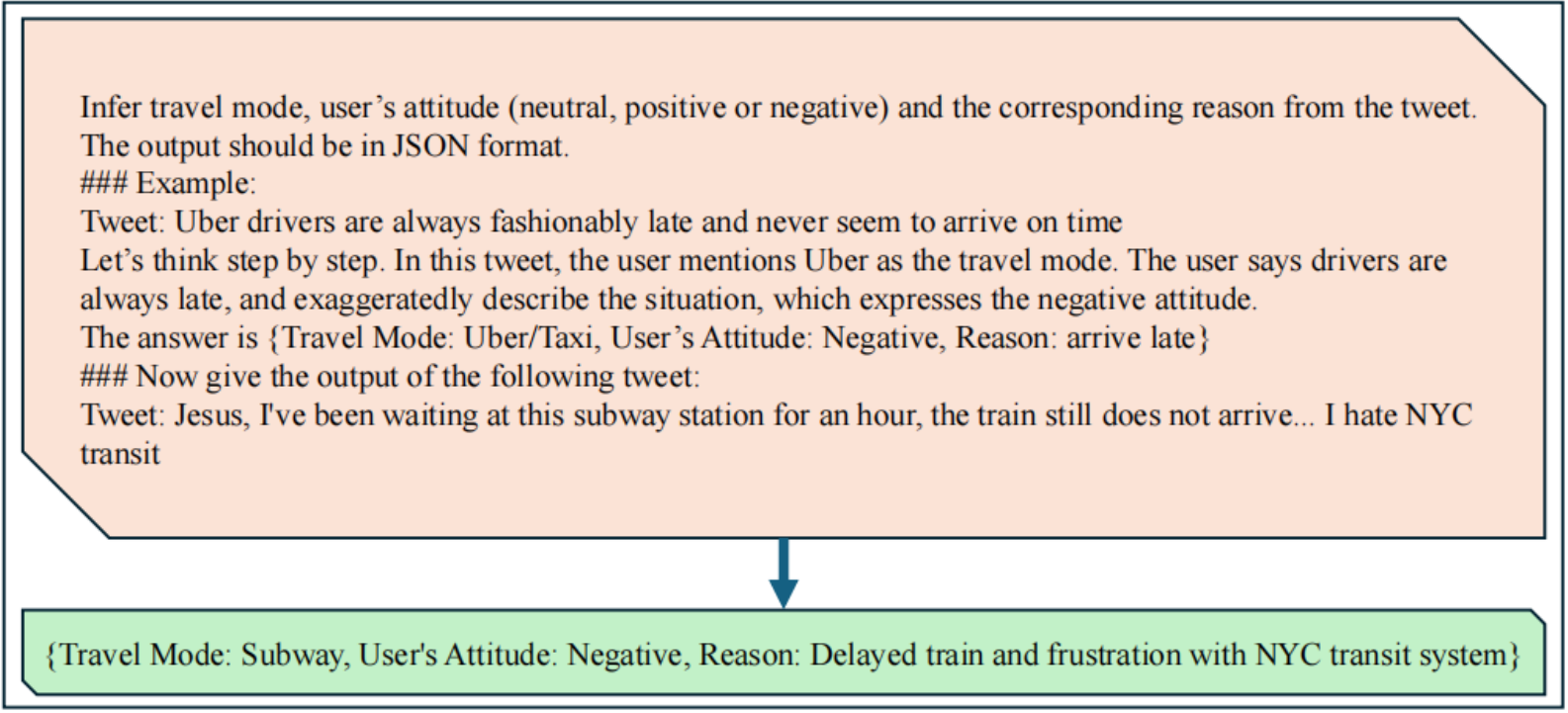}
    \caption{Chain-of-thought}
\label{fig:prompting:cot}
\end{subfigure}
\hfill
\begin{subfigure}[b]{1\columnwidth}\centering
    \includegraphics[width=\linewidth]{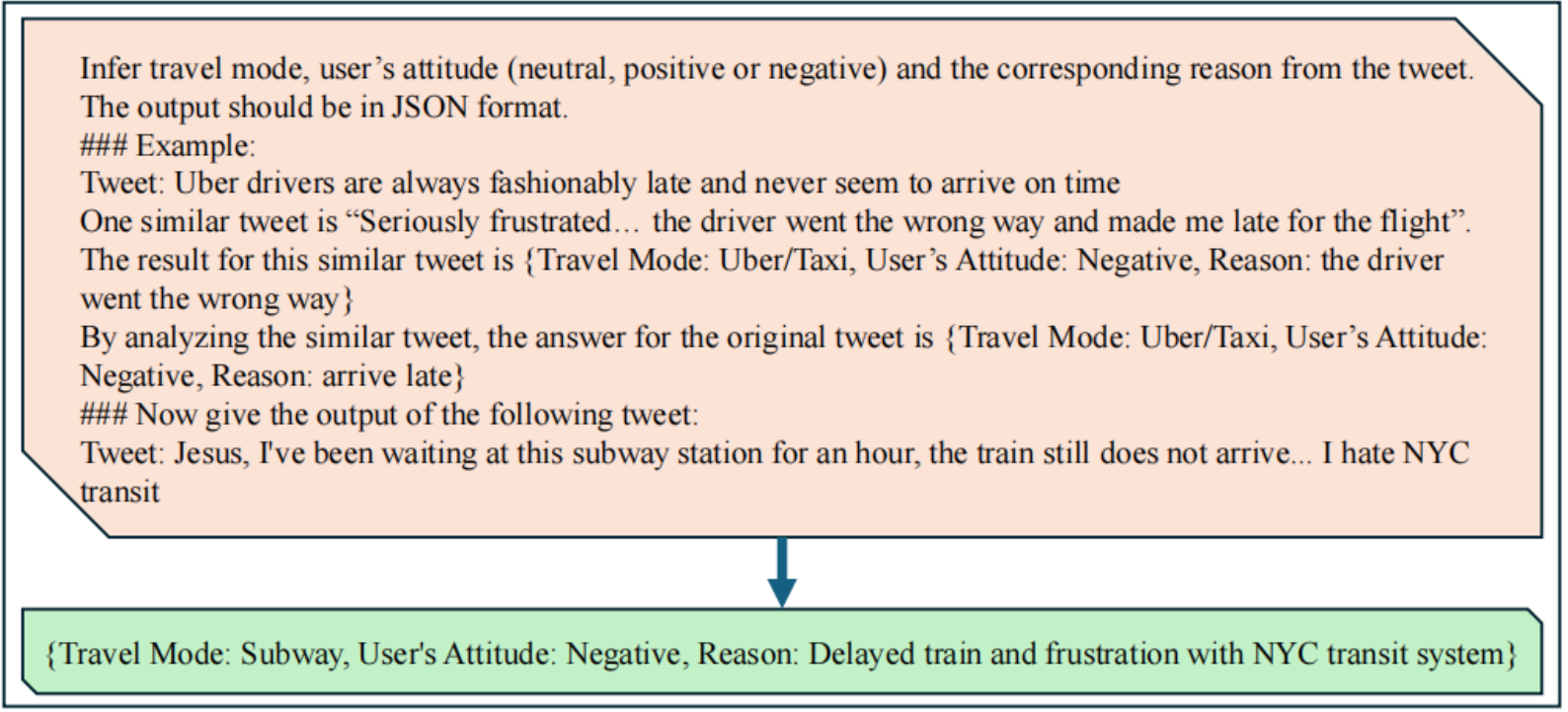}
    \caption{Analogical Prompting} 
\label{fig:prompting:analogical}
\end{subfigure}
}
\caption{Visualization of different prompting engineering methods.}
\label{fig:prompting}
\vspace{-6.5 mm}
\end{figure*}

\vspace{-2.0 mm}
\section{Methodology} \label{sec:methodology}
\vspace{-1.0 mm}


We employ a similar pre-processing procedure for the collected tweets, which is consistent with the prior research \cite{maghrebi2016transportation,chen2023sentiment}.
The overall structure of the proposed framework is illustrated in \Cref{fig:overall_structure}. There are two primary LLM agents: the reasoner and the verifier. 
For a given tweet, the reasoner predicts the corresponding travel mode, assesses the sentiment, and conducts a reasoning check.
Subsequently, the verifier examines and confirms the validity of these predictions. 

\vspace{-1.0 mm}
\subsection{Large Language Models (LLMs)}
\vspace{-1.0 mm}

The mainly chosen LLMs in this paper include GPT-3.5 \cite{achiam2023gpt}, Llama2 \cite{touvron2023llama}, and Mistral \cite{jiang2023mistral}.
All of them are advanced Transformer-based models with over billions of parameters and pre-trained on a vast corpus of text data. The considerable size of the model and diverse training dataset enable LLMs like GPT-3.5 to exhibit remarkable capabilities, e.g., zero-shot learning \cite{wei2021finetuned}, and solving problems with step-by-step reasoning \cite{kojima2022large}.

The fundamental building block of the Transformer architecture is the attention mechanism \cite{vaswani2017attention,ruan2024infostgcan}. 
Specifically, with notations consistent with previous studies \cite{vaswani2017attention,choromanski2020rethinking}, $L$ denotes the length of the input sequence of tokens. Attention projects the input into three different vectors: queries $\boldsymbol{\mathbf{Q}}$, keys $\boldsymbol{\mathbf{K}}$ and values $\boldsymbol{\mathbf{V}}$ \cite{ruan2024s2e}. 
For the bidirectional dot-product attention $\operatorname{Attn}_{\leftrightarrow}$, the outcome can be computed as follows:
\vspace{-1.0 mm}
\begin{align} \label{eqn:bi_attn}
\begin{split}
    & \operatorname{Attn}_{\leftrightarrow}(\mathbf{Q}, \mathbf{K}, \mathbf{V})=\mathbf{D}^{-1} \mathbf{A} \mathbf{V}, \\ 
    & \mathbf{A}=\exp \left(\mathbf{Q} \mathbf{K}^{\top} / \sqrt{d}\right), \quad \mathbf{D}=\operatorname{diag}\left(\mathbf{A} \mathbf{1}_{L}\right)
\end{split}
\end{align}
where $\exp(\cdot)$ signifies the element-wise exponential function,
$\mathbf{K}^{\top}$ represents the transpose of $\mathbf{K}$.
$\operatorname{diag}(\cdot)$ extracts the diagonal elements of a matrix, and $\mathbf{1}_{L}$ denotes a vector full of ones. 
Another significant type is unidirectional $\operatorname{Att}_{\rightarrow}$:
\begin{align} \label{eqn:uni_attn}
\begin{split}
    & \operatorname{Att}_{\rightarrow}(\mathbf{Q}, \mathbf{K}, \mathbf{V})=\widetilde{\mathbf{D}}^{-1} \widetilde{\mathbf{A}} \mathbf{V}, \quad \widetilde{\mathbf{D}}=\operatorname{diag}\left(\widetilde{\mathbf{A}} \mathbf{1}_{L}\right), \\ & \quad \widetilde{\mathbf{A}} = \operatorname{tril}(\mathbf{A}), \quad \mathbf{A}=\exp \left(\mathbf{Q} \mathbf{K}^{\top} / \sqrt{d}\right) \\
\end{split}
\end{align}
where $\operatorname{tril}(\cdot)$ returns the triangular part of the input matrix with the diagonal. Unidirectional dot-product attention is important for autoregressive generative
modelling \cite{brown2020language}.

The primary difference between the bidirectional dot-product attention (\Cref{eqn:bi_attn}) and unidirectional dot-product attention (\Cref{eqn:uni_attn}) lies in the application of $\operatorname{tril}(\cdot)$. In unidirectional dot-product attention, each position in the sequence is only allowed to attend to the preceding positions and itself. The function $\operatorname{tril}(\cdot)$ helps to ensure that future positions are masked, and therefore, not attended to during the attention computation.

\vspace{-1.0 mm}
\subsection{Prompting Engineering Methods} \label{sec:prompting_eng_methods}
\vspace{-1.0 mm}

Briefly speaking, a prompt constitutes the input provided to LLMs \cite{brown2020language}. The practice of designing language queries to guide the model’s outputs towards specific goals is commonly known as prompt engineering \cite{zhao2023survey}. 
The syntax and semantics of a prompt can significantly affect a model’s response.

Specifically, 
we compare the following prompting methods: instruction-following \cite{raffel2020exploring}, in-context learning \cite{brown2020language}, chain-of-thought \cite{wei2022chain}, and analogical prompting \cite{yasunaga2024large}. To highlight their unique characteristics and differences, examples are provided in \Cref{fig:prompting}.





\begin{itemize}
    \item Instruction-following \cite{raffel2020exploring} involves direct commands that guide the response generation of the language model.

    \item In-Context Learning (ICL) \cite{brown2020language} is a paradigm where LLM acquire the capability to perform new tasks through inference alone, without the need for updating its parameters.
    
    \item Chain-of-thought prompting \cite{wei2022chain} encourages the model to articulate intermediate steps towards the solution, fostering a more transparent reasoning process, e.g., ``Let's think step by step".

    \item Inspired from human beings utilizing past experiences to solve new problems, Analogical prompting \cite{yasunaga2024large} enables language models to self-generate relevant examples or knowledge within a specific context. 
\end{itemize}

\begin{figure*}[!ht]
\centering
  \includegraphics[width = 0.92 \linewidth]{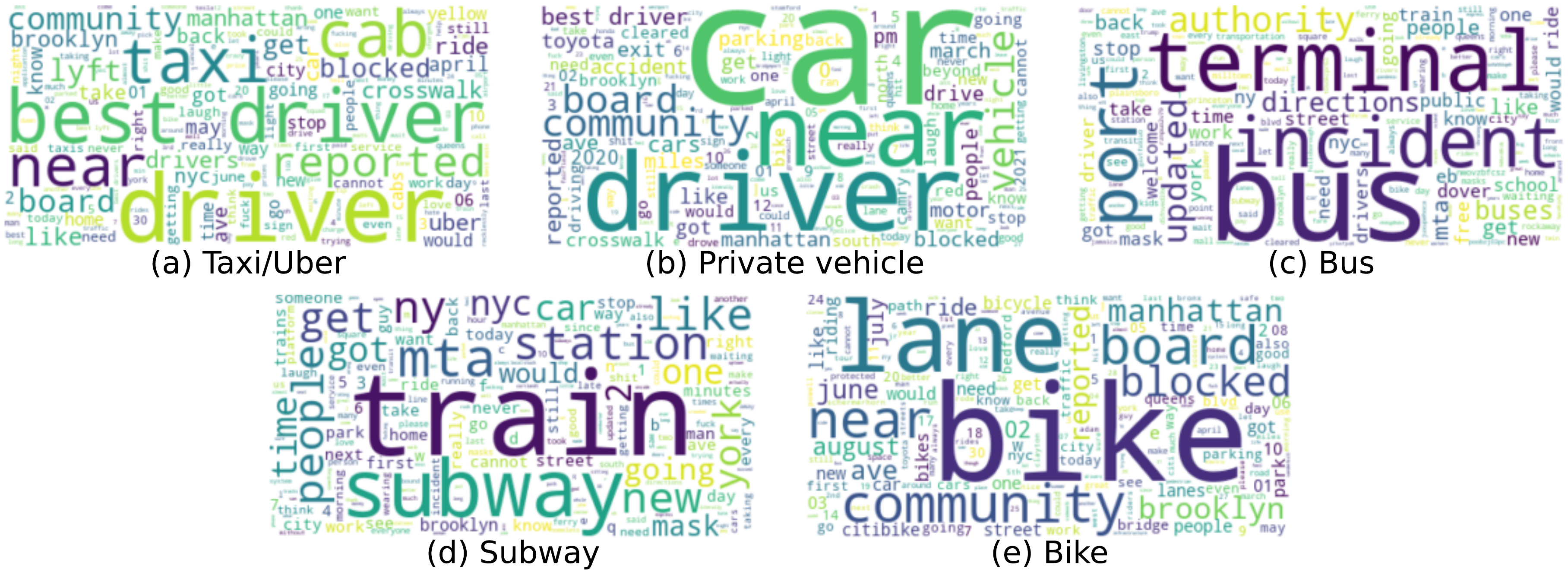}
  \vspace{-1. mm}
  \caption{The wordcloud maps for different travel modes predicted by the reasoner.}
  \vspace{-5. mm}
  \label{fig:wordcloud}
\end{figure*}

\vspace{-2.0 mm}
\section{Results and Discussion} \label{sec:results}
\vspace{-1.0 mm}

In this section, we first identify the optimal LLM and prompt engineering technique for the reasoner. Based on the chosen methodologies, we conduct a detailed analysis of the distribution patterns of travel modes, and the sentiments associated with each. Ultimately, we summarize the primary factors contributing to dissatisfaction and propose targeted strategic recommendations to ameliorate these concerns.

Our evaluation integrates human assessment and LLM verification to gauge performance.
Specifically, each LLM is provided with identical input prompts to generate responses. These output responses are then assessed for coherence and relevance by both human evaluators and LLMs. In particular, human evaluators are asked to score the responses based on specific criteria, e.g., the correctness of the travel mode or the sentiment. Similarly, inspired by \cite{peng2023instruction}, the LLM verifier also evaluates such aspects, utilizing GPT-4 \cite{achiam2023gpt}. 
Combining human assessment and LLM verification offers a more robust approach to evaluate language models.
Scores are normalized on a scale from 0 to 1, where higher values indicate superior performance. For each LLM, the average score across the test dataset is computed to determine its performance efficacy.

\vspace{-1.0 mm}
\subsection{Ablation Study on Different LLMs and Prompting Engineering Methods}
\vspace{-1.0 mm}

In order to choose the optimal LLM for the reasoner, we conduct an ablation study. Specifically, we examine the following  representative LLMs: GPT-3.5 \cite{achiam2023gpt}, Llama2-7B \cite{touvron2023llama}, and Mistral-7B \cite{jiang2023mistral}.

\begin{table}[ht!]
\vspace{-2. mm}
    \centering
    \caption{The evaluation scores for different LLMs.}
    \begin{tabular}{ c | c  c | c c} 
     \thickhline
     Models & \multicolumn{2}{c|}{Human Verification Score} & \multicolumn{2}{c}{LLM Verification Score} \\  \cline{2-5}
      & Travel Mode & Sentiment & Travel Mode & Sentiment \\ 
     \hline
     GPT-3.5 & \textbf{0.82} & \textbf{0.75} &  \textbf{0.96} & \textbf{0.79} \\
     Llama2-7B & 0.74 & 0.68 & 0.77 & 0.59 \\
     Mistral-7B & 0.73 & 0.66 & 0.87 & 0.69 \\
     \thickhline
    \end{tabular}
    \label{table:1}
    \vspace{-4 mm}
\end{table}

The results, detailed in \Cref{table:1}, show that GPT-3.5 consistently achieved the highest average scores across both human verification and LLM verification. Generally, human verification scores are generated based on hundreds of tweets, while LLM verification scores are generated based on thousands of tweets.
Specifically, consider a real-world tweet ``sorry to ask is being miserable a criteria to be employed by the mta? almost every mta employee is miserable and angry". 
The response for GPT-3.5 is: ``The travel mode related to the tweet is likely Metro, because of MTA. The sentiment expressed in the tweet is negative, as the user is expressing frustration and dissatisfaction with the attitude of MTA employees." 
In contrast, Llama2-7B generates a response: ``Travel mode: Walking, Sentiment: Negative". 
The response of Mistral-7B begins with ``The travel mode related to the tweet is not explicitly stated".
Consequently, we choose GPT-3.5 as the default model for the reasoner.

\begin{table}[ht!]
\vspace{-1. mm}
\centering
\caption{The evaluation scores for different prompting engineering methods.}
\resizebox{\columnwidth}{!}{
\begin{tabular}{ c | c  c | c c} 
 \thickhline
 Models & \multicolumn{2}{c|}{Human Verification Score} & \multicolumn{2}{c}{LLM Verification Score} \\ \cline{2-5}
  & Travel Mode & Sentiment & Travel Mode & Sentiment \\
 \hline 
 Instruction-following & 0.83 & 0.72 &  0.80 & 0.76 \\
 ICL & \textbf{0.84} & \textbf{0.73} & \textbf{0.93} & \textbf{0.81} \\
 Chain-of-thought & 0.77 & 0.68 & 0.87 & 0.77 \\
Analogical & 0.70 & 0.61 & 0.64 & 0.50 \\
 \thickhline
\end{tabular}
}
\vspace{-3. mm}
\label{table:2}
\end{table}

As shown in \Cref{table:2}, in-context learning has the best performance among all prompting methods (\Cref{fig:prompting}). Consequently, we choose in-context learning as the default prompting method for the reasoner.

\vspace{-1.0 mm}
\subsection{Travel Mode and Sentiment Analysis} \label{sec:travel_mode_and_sentiment}

To further validate the predictive accuracy of our framework, we examined the most frequent words in tweets associated with each travel mode predicted by the reasoner.
\Cref{fig:wordcloud} visualizes the frequent words by a series of word cloud maps, providing insights into the linguistic patterns related to different modes of travel.
Specifically, \Cref{fig:wordcloud}(a) highlights terms like ``driver", ``best driver", ``taxi", and ``cab", reflecting common discussions related to taxi or Uber services. 
\Cref{fig:wordcloud}(b) shows the word cloud for private vehicle. 
\Cref{fig:wordcloud}(c) represents bus, logically encompassing words such as ``bus", ``terminal", ``port", and ``public".
\Cref{fig:wordcloud}(d) shows the word cloud for subway, including terms like ``train", ``subway", ``station", ``mta". 
\Cref{fig:wordcloud}(e) shows the bike-related word cloud, including terms like ``bike", ``lane", ``path", ``ride" underscoring the relevance of these words to bike.

\begin{figure}[!ht]
\centering
  \includegraphics[width = \linewidth]{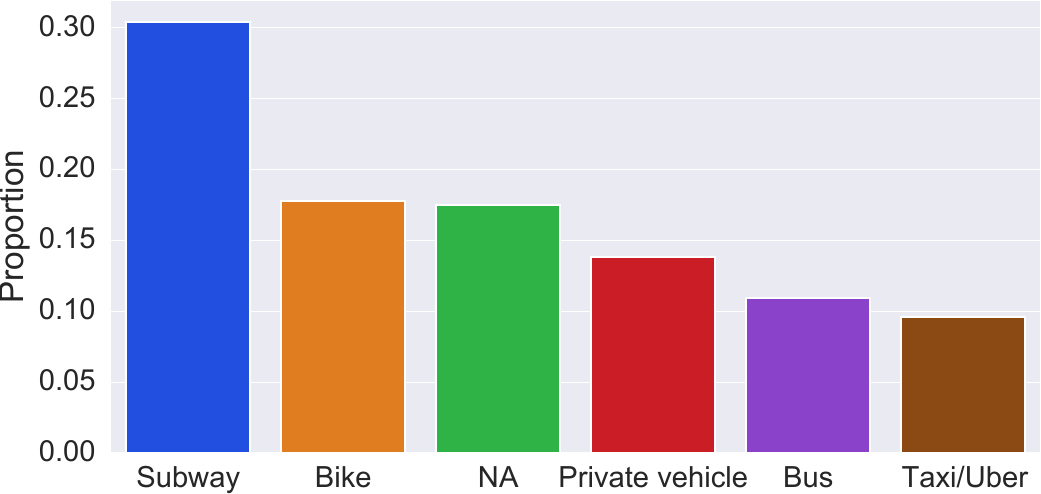}
  \caption{The proportion of each travel mode extracted from collected Twitter data.}
  \vspace{-7 mm}
  \label{fig:distribution_hist}
\end{figure}

\Cref{fig:distribution_hist} illustrates the distribution of travel modes in NYC, based on the collected Twitter data.
Remarkably, the subway/metro emerges as the predominant mode of transportation, which likely reflects the efficiency of NYC’s extensive metro system and the flexibility it offers to commuters. 
Following the subway, bike ranks as the second most popular mode. This preference can be understood in the context of the city's congested traffic conditions, which often make cycling a faster and more convenient option.

"NA" appears third in the ranking, indicating instances where the context of keywords such as ``subway" could lead to ambiguous interpretations. 
For example, when collecting tweets using the keyword ``subway", it may refer to the metro system or the restaurant. 
An illustrative tweet like ``I like the sandwich at Subway in NYC" is actually not related to any travel mode, as the term ``subway" in this context is highly likely to refer to the restaurant chain. Such ambiguity could contribute to the high incidence of `NA' as a travel mode.

Subsequently, private vehicles are the fourth most common mode, followed by buses and then taxis/Uber, which rank sixth. 
In urban cities, such as NYC, the metro system's extensive network and convenience make it extremely popular, and biking might be preferred due to faster navigation through congested streets. In contrast, private vehicles, buses, and taxis/Uber might be less preferred, as they face challenges like parking scarcity, congestion, and higher costs.

\begin{figure}[!ht]
\centering
  \includegraphics[width = \linewidth]{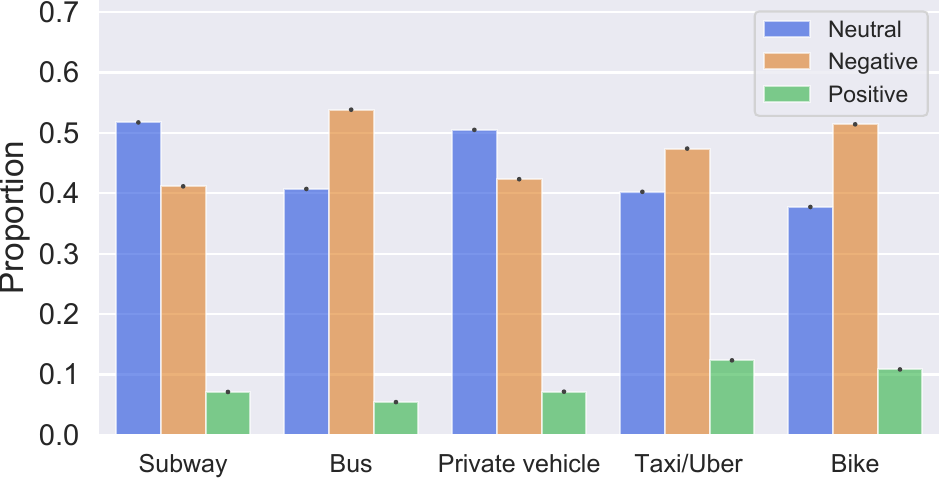}
  \vspace{-4. mm}
  \caption{Users' attitudes (Neutral/Negative/Positive) towards different travel modes.}
  \vspace{-4. mm}
 \label{fig:satisfaction_freq_bar}
\end{figure}

\Cref{fig:satisfaction_freq_bar} shows the proportion of user attitudes (Neutral/Negative/Positive) towards different travel modes.
The results indicate that most travel modes exhibit a higher proportion of negative responses compared to positive ones. This trend aligns with existing research \cite{gregoire2015managing,yen2016factors}, suggesting that individuals with negative experiences are more likely to share their grievances on social media platforms.
Notably, Taxi/Uber and bicycles demonstrate slightly higher levels of user satisfaction comparing to other modes of travel.

\begin{figure}[!htp]
\centering
\hfill
\begin{subfigure}{0.46\linewidth}
\includegraphics[width=\textwidth]{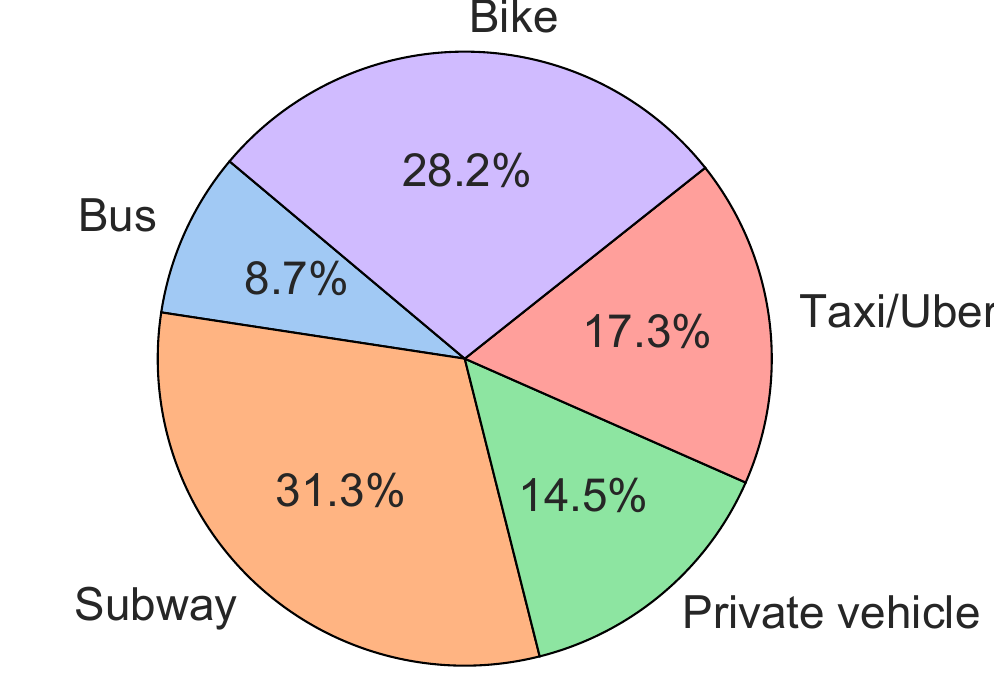}
\caption{Proportion of travel modes when the sentiment is \emph{positive}.}
\label{fig:piechart_Positive}
\end{subfigure}\hfill
\begin{subfigure}{0.5\linewidth}
\includegraphics[width=\textwidth]{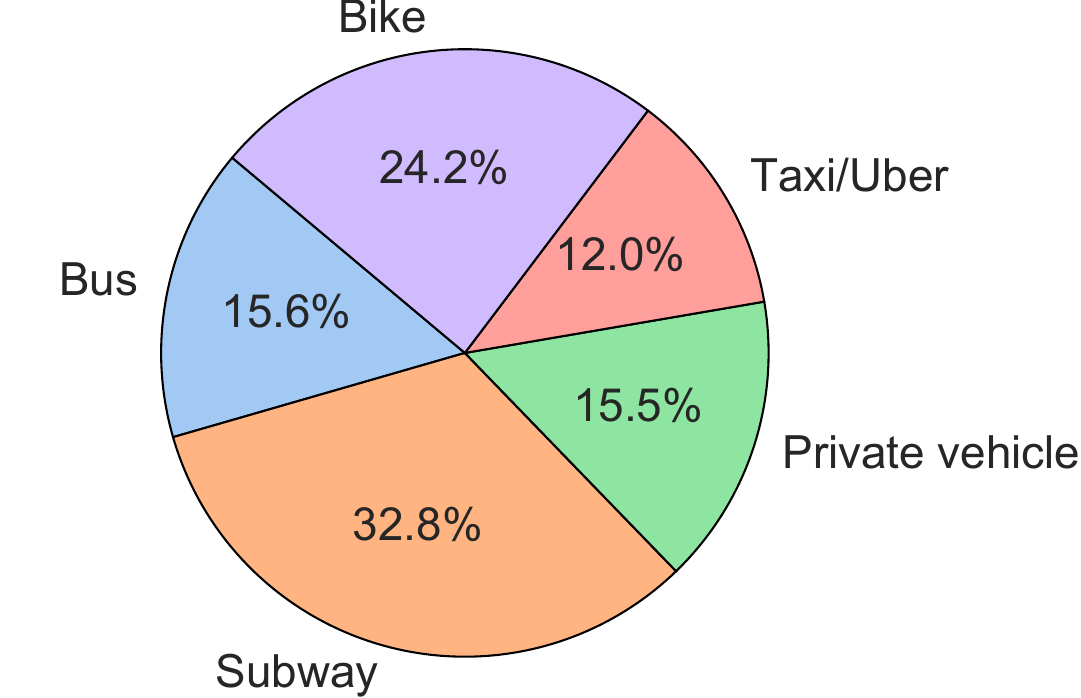}
\caption{Proportion of travel modes when the sentiment is \emph{negative}.}
\label{fig:piechart_Negative}
\end{subfigure}\par
\vskip\floatsep
\vspace{-6. mm}
\caption{Sentiment distributions across different travel modes: positive, and negative.}
\vspace{-7. mm}
\end{figure}

\Cref{fig:piechart_Positive}, \Cref{fig:piechart_Negative},  depict the travel modes distribution across different sentiments, specifically highlighting positive and negative, respectively. 
As previously discussed, while Taxi/Uber tweets are fewer, these modes exhibit marginally higher satisfaction levels.
The subway, having the highest number of tweets, also shows the largest share in each sentiment category. 

\vspace{-1.0 mm}
\subsection{Major Reasons of Dissatisfaction}
\vspace{-1.0 mm}

\begin{figure}[!ht]
\centering
    \vspace{-2. mm}
  \includegraphics[width = \linewidth]{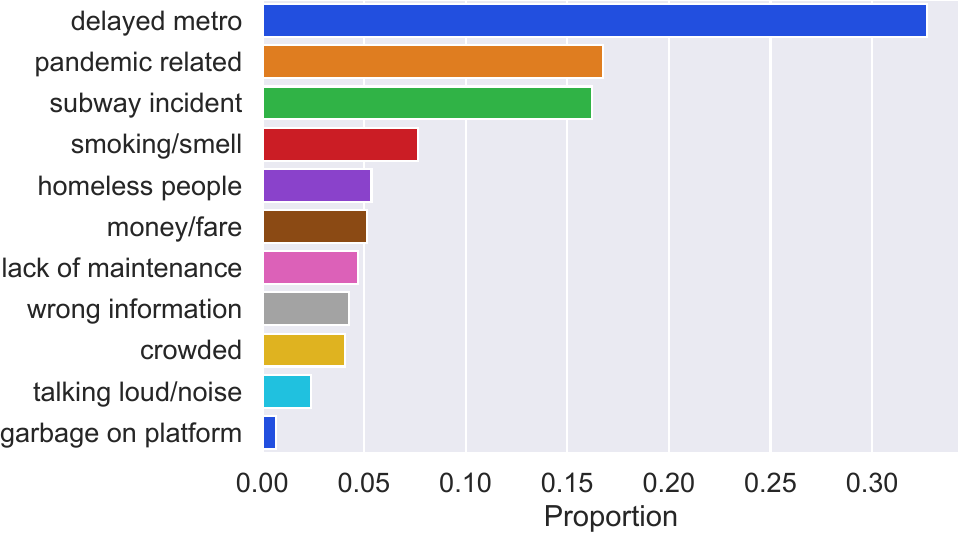}
  \vspace{-5. mm}
  \caption{Representative factors causing \textit{subway} dissatisfaction.}
  \vspace{-4 mm}
  \label{fig:metro_reason_prop}
\end{figure}

\begin{figure}[!ht]
\centering
    \vspace{-2. mm}
  \includegraphics[width = \linewidth]{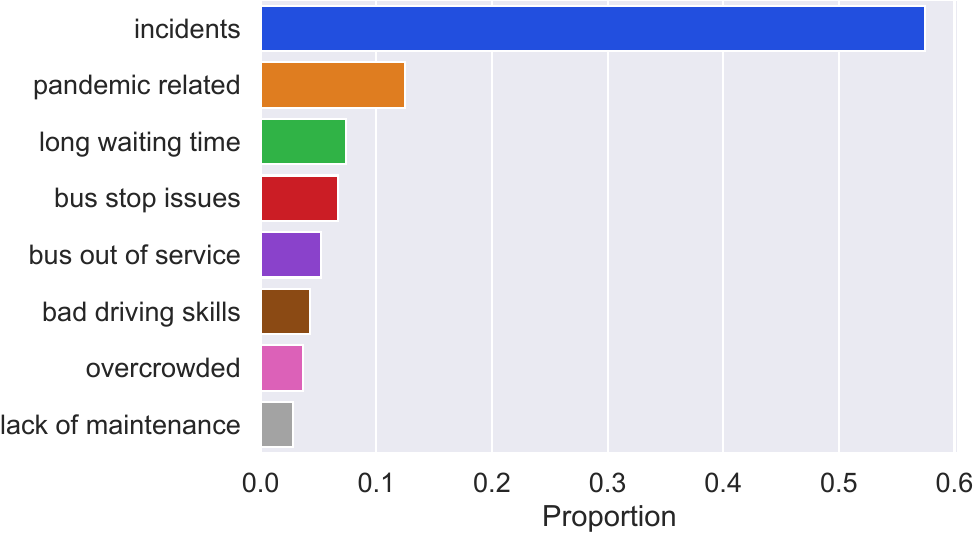}
  \vspace{-5. mm}
  \caption{Representative factors causing \textit{bus} dissatisfaction.}
  \vspace{-3 mm}
  \label{fig:bus_reason_prop}
\end{figure}

\begin{figure}[!ht]
\centering
    \vspace{-2. mm}
  \includegraphics[width = \linewidth]{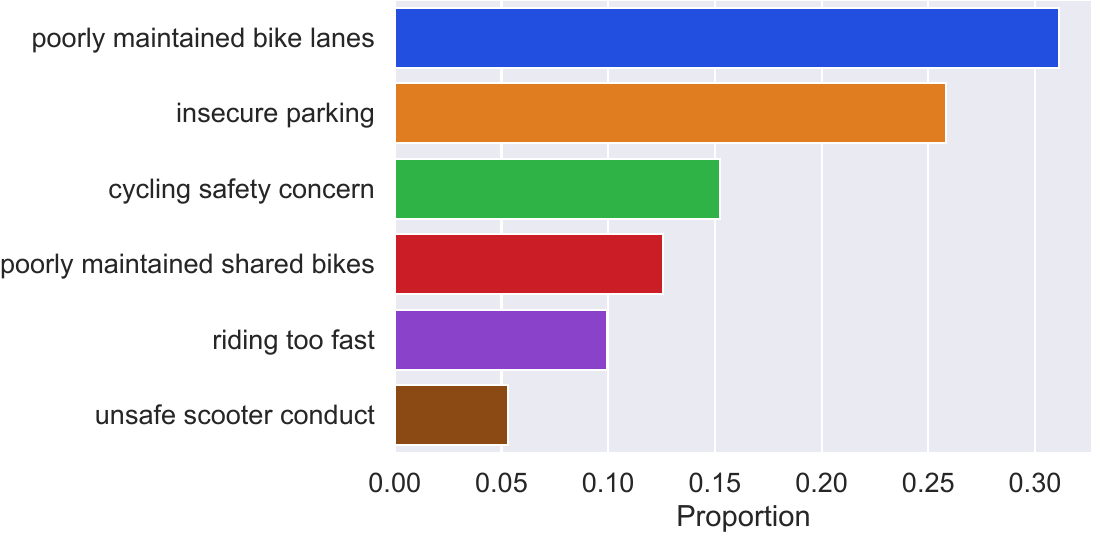}
  \vspace{-5. mm}
  \caption{Representative factors causing \textit{bike} dissatisfaction.}
  \vspace{-3 mm}
  \label{fig:bike_reason_prop}
\end{figure}

\begin{figure}[!ht]
\centering
  \includegraphics[width = \linewidth]{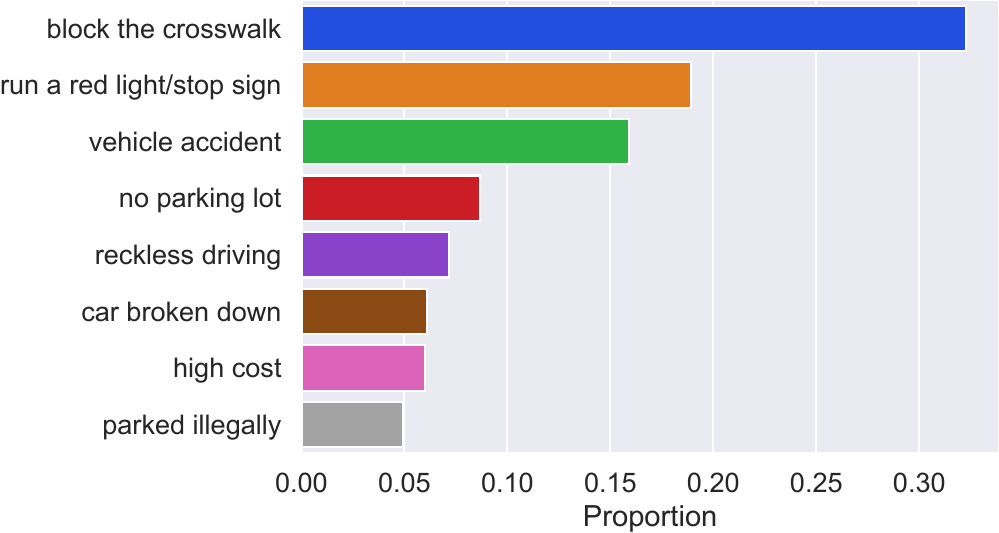}
  \vspace{-5. mm}
  \caption{Representative factors causing dissatisfaction among \textit{taxi/Uber} and \textit{private vehicle}.}
  \vspace{-5. mm}
\label{fig:vehicle_reason_prop}
\end{figure}

To elucidate the representative factors contributing to dissatisfaction among different travel modes, we conduct a comprehensive analysis.
\Cref{fig:metro_reason_prop}, \Cref{fig:bus_reason_prop}, \Cref{fig:bike_reason_prop}, and \Cref{fig:vehicle_reason_prop} illustrate the primary reasons for negative feedback specific to different modes.
In these figures, the complaints are ranked from most to least frequent.

To comprehend the underlying factors contributing to the dissatisfaction of nearly 40\% of subway users, we analyzed the data and identify the most prevalent complaints, as depicted in \Cref{fig:metro_reason_prop}.
The analysis reveals that the primary complaint is delays and long waiting time. The second most common complaint was inadequate COVID-19 safety measures, including improper mask usage and insufficient physical distancing. The third issue involves incidents on the subway, including racist and harassment incidents. Additionally, users reported problems with smoking, odors, homelessness, fare concerns, maintenance shortcomings, misinformation, noise, and litter.
Based on our findings, we recommend enhancing timeliness and reliability, enforcing health protocols during the pandemic, improving security, and addressing environmental and operational concerns.


\Cref{fig:bus_reason_prop} presents the analysis for dissatisfaction for bus, highlighting some issues, e.g., bus incidents, pandemic-related concerns, long waiting time, problems of bus stops, and so on.
To enhance service quality, we recommend enhancing bus drivers' training and professionalism, improving service reliability, ensuring strict adherence to health protocols, increasing maintenance frequency for better quality.

\Cref{fig:bike_reason_prop} depicts the major dissatisfaction factors for bike, including inadequate maintenance of bike lanes, lack of secure parking, safety issues while cycling, substandard conditions of shared bicycles, excessive speed by some cyclists, and unsafe scooter behaviors. To address these issues, we advocate for enhancing the quality of bike lanes, developing secure bicycle parking facilities, and improving the standards of shared bicycles.

As shown in \Cref{fig:vehicle_reason_prop}, we analyze dissatisfaction factors related to taxi/Uber and private vehicles together, due to shared vehicular concerns. 
The key issues identified include both vehicle-related problems, such as obstructions of crosswalks, violations of traffic signals including running red lights or stop signs, accidents, reckless driving, illegal parking; and user-related concerns, for example, the scarcity of parking spaces and the high costs of ride-sourcing services.
To mitigate these problems, we recommend increasing penalties for traffic violations such as obstructing crosswalks or non-compliance with traffic signals. Additionally, we recommend the expansion of parking infrastructure and a strategic reduction in ride-sourcing service costs.

\vspace{-1.0 mm}
\section{Conclusion} \label{sec:conclusion}
\vspace{-1.0 mm}

In this work, we introduce a novel LLM-based framework to analyze and extract individuals' travel mode choices from Twitter data, without the need of manual annotations. 
Our framework consists of the `reasoner' that predicts travel modes and sentiments, and the `verifier' that validates these predictions.
We evaluate various LLMs and prompting strategies, and find that GPT-3.5 surpasses Llama2-7B and Mistral-7B. Moreover, our results show that in-context learning is particularly effective for the reasoner. 
Given that the dataset is mainly collected in NYC, subway/metro emerged as the most frequent travel mode, followed by bikes, private vehicles, buses, and taxis/Uber.
Furthermore, our analysis suggests that individuals with negative experiences are more likely to express their dissatisfaction on social media. Accordingly, we identify the major causes of discontent for different modes and propose several recommendations to address these issues.

\vspace{-2.0 mm}








{\small
\bibliographystyle{IEEEtran}
\bibliography{references}
}

\end{document}